\documentclass{article}

\usepackage{arxiv}

\usepackage[utf8]{inputenc} 
\usepackage[T1]{fontenc}    
\usepackage{hyperref}       
\usepackage{url}            
\usepackage{booktabs}       
\usepackage{amsfonts}       
\usepackage{nicefrac}       
\usepackage{microtype}      
\usepackage{lipsum}		
\usepackage{graphicx}
\usepackage{float} 
\usepackage{subcaption}
\usepackage[square,numbers]{natbib} 
\usepackage{doi}
\usepackage{graphicx}
\usepackage{subcaption}
\usepackage[export]{adjustbox}
\usepackage{wrapfig}
\usepackage{siunitx}
\usepackage{xcolor, color, soul}
\usepackage{multirow}
\title{Evaluating self-attention interpretability through human-grounded experimental protocol}

\author{ \href{https://myedb.edite-de-paris.fr/Fiche/41003/}{\includegraphics[scale=0.06]{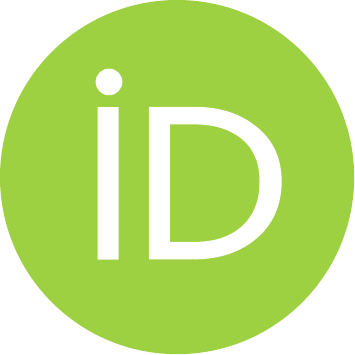}\hspace{1mm}Milan Bhan} \\
	Ekimetrics\\
	Sorbonne University\\
	Paris \\
	\texttt{milan.bhan@ekimetrics.com} \\
	\And
	\href{https://www.linkedin.com/in/nina-achache/}{\includegraphics[scale=0.06]{orcid.pdf}\hspace{1mm}Nina Achache} \\
	Ekimetrics\\
	Paris \\
	\texttt{nina.achache@ekimetrics.com} \\
     \And
    	\href{https://www.linkedin.com/in/legrand-victor/}{\includegraphics[scale=0.06]{orcid.pdf}\hspace{1mm}Victor Legrand} \\
    	Ekimetrics\\
    	Paris \\
    	\texttt{victor.legrand@ekimetrics.com} \\
     \And
    	\href{https://www.linkedin.com/in/annabelleblangero/?originalSubdomain=fr}{\includegraphics[scale=0.06]{orcid.pdf}\hspace{1mm}Annabelle Blangero} \\
        Aix-Marseille University \\
    	Ekimetrics\\
    	Paris \\
    	\texttt{annabelle.blangero@ekimetrics.com} \\
    \And
    	\href{https://www.linkedin.com/in/nicolas-chesneau-0416b94/}{\includegraphics[scale=0.06]{orcid.pdf}\hspace{1mm}Nicolas Chesneau} \\
    	Ekimetrics\\
    	Paris \\
    	\texttt{nicolas.chesneau@ekimetrics.com} \\
}

\renewcommand{\shorttitle}{Bhan et al.}

\hypersetup{
pdftitle={A template for the arxiv style},
pdfsubject={q-bio.NC, q-bio.QM},
pdfauthor={David S.~Hippocampus, Elias D.~Striatum},
pdfkeywords={Interpretability, Explainability, Attention, Semantic congruency, Natural language processing},
}

\begin{document}
\maketitle

\begin{abstract}
Attention mechanisms have played a crucial role in the development of complex architectures such as Transformers in natural language processing. However, Transformers remain hard to interpret and are considered as black-boxes. This paper aims to assess how attention coefficients from Transformers can help in providing interpretability.  A new attention-based interpretability method called CLaSsification-Attention (CLS-A) is proposed. CLS-A computes an interpretability score for each word based on the attention coefficient distribution related to the part specific to the classification task within the Transformer architecture. A human-grounded experiment is conducted to evaluate and compare CLS-A to other interpretability methods. The experimental protocol relies on the capacity of an interpretability method to provide explanation in line with human reasoning. Experiment design includes measuring reaction times and correct response rates by human subjects. CLS-A performs comparably to usual interpretability methods regarding average participant reaction time and accuracy. The lower computational cost of CLS-A compared to other interpretability methods and its availability by design within the classifier make it particularly interesting. Data analysis also highlights the link between the probability score of a classifier prediction and adequate explanations. Finally, our work confirms the relevancy of the use of CLS-A and shows to which extent self-attention contains rich information to explain Transformer classifiers.
\end{abstract}


\section{Introduction}
The field of machine learning (ML) has witnessed great advance in recent years. ML algorithms have achieved high levels of performance in a wide variety of tasks thanks to their rapid development and improving complexity. The complexity of these new models has led to an increasing difficulty in understanding and interpreting their predictions. The field of eXplainable Artificial Intelligence (XAI) has emerged to overcome this lack of transparency by developing methods of "interpretability" or "explicability". Most commonly used interpretability methods are time consuming and usually based on strong hypothesis such as features’ independence and linear approximation~\cite{poursabzi2021manipulating}. The explanation provided can differ significantly from one method to another.
Natural language processing (NLP) has been particularly concerned by recent breakthroughs in ML with the development and democratization of Transformer-type models \cite{Attention}. Attention mechanism \cite{Neural} is a crucial component in Transformer architecture, enabling models to focus on specific parts of the input text. However, the interpretability of attention remains questionable~\cite{notnotXP, notXP}. The plurality of interpretability methods and the differences in the resulting explanations raise the question of their assessment and comparison.

This paper aims to assess the interpretability of attention mechanisms in Transformers models. A new method which we call CLaSsification-Attention (CLS-A) is built based on attention weights from Transformers. Additionally, this work aims to illustrate how human-based experimental protocols can be good alternatives for interpretability comparison.
We first introduce key notions of Transformers architecture and interpretability. We then introduce our new attention-based method CLS-A and the experimental protocol used to assess its interpretability. Finally, we analyze the experimental results which validate the CLS-A's interpretability.
Our work adds to the literature questioning the interpretability of attention coefficients in recent Transformer models. CLS-A seems particularly appropriate to explain attention-based classifiers in NLP since we validate its interpretability and because attention is available at no cost within the model architecture.

\section{Background \& related work}
\label{sec:headings}
This section introduces some key notions about Transformers architectures and interpretability as well as an overview of existing approaches to compare XAI methods.  
\subsection{Background}
\paragraph{Transformers, attention and BERT} In NLP, Transformer-like models have achieved high levels of performance in a variety of tasks, such as text summarization, question answering or named entity recognition. These models are particularly complex, with a number of parameters that can exceed the billion \cite{DistilBERT}. The Transformer architecture is based on multi-head self-attention mechanisms, aiming at making learning more efficient \cite{Neural} by encoding the relations between words. The model attends to different parts of the input in parallel, using multiple self-attention heads. A self-attention head takes as input a triplet ($Q$, $K$, $V$) and outputs a representation as formalized in the following formula :
\begin{equation}
\label{eqn:attention}
\texttt{Attention}(Q,K,V) = \texttt{softmax}(\frac{QK^{T}}{\sqrt{d_{k}}})V
\end{equation}
Where:
\begin{itemize}
    \item $Q$ (query) is a matrix that represents the input in which the attention mechanism focuses on. 
    \item $K$ (key) is a matrix that represents the different elements in the input that the attention mechanism can attend to.
    \item $V$ (value) is a matrix that represents the output of the attention mechanism.
    \item $d_{k}$ is the dimension of matrices $Q$ and $V$ and allows to stabilize the model during the training phase
\end{itemize}
Hence, each head has its own set of parameters, allowing the model to learn different types of attention patterns. The attentions resulting from each of the heads are then concatenated and projected on a dense layer. 

Bidirectional Encoder Representations from Transformers (BERT) models \cite{devlin_bert_2019} is a stack of $n$ encoders from the Transformer architecture. Each BERT layer contains $h$ attention heads with its own set of weights, which have been learned during training. These weights determine how the model will attend to different parts of the input when making a prediction. In this way, words are related to each other even in the case of long-term dependency. BERT has been widely adopted and has achieved state-of-the-art performance on a variety of benchmarks such as GLUE\cite{wang-etal-2018-glue} for natural language understanding (NLU). 

One of the key features of BERT is its bidirectional nature. Unlike previous models that were only trained to look \textit{before} or \textit{after} a word, BERT is trained to look \textit{before} and \textit{after} a word at the same time. This allows BERT to understand the full context of a word and improve its performance on NLU tasks. Moreover, BERT has several advantages such as its scalability, the fact that it can be parallelized\cite{deep_mind_transfo} and that it can compute longer sequences. The \textit{CLS} token (short for "classification") in BERT is a special token added to the beginning of the input text that the model uses as a representation of the entire input sentence or document. The final hidden state of the \textit{CLS} token, which is a fixed-size vector, is typically used as the input to a classifier or other downstream task. This allows BERT to take into account information from the entire sentence or document when making a prediction.

\paragraph{Local feature importance.} There are several ways to interpret black-box systems such as BERT models \cite{guidotti_counterfactual_2022}. One of the main approaches consists in computing local feature importance \cite{molnar_interpretable_2020}. When the model to explain is a classifier, the purpose is to compute contribution coefficients to the probability score of the predicted class. It can be done by considering ML models as black-boxes and explaining their predictions \textit{ex post}, without referring to their inherent parameters. This kind of approach is called \textit{post-hoc model-agnostic}\cite{MLint_Survey}. Another way to compute local feature importance is to use the information contained within the model architectures\cite{Axio, Importance}, which can significantly reduce the computation time. However, it makes the method rely strongly on the model it tries to explain and makes its use less flexible. This kind of approaches is referred to as \textit{post-hoc model-specific}. The large number of local feature importance methods can make it difficult to choose the most appropriate one \cite{SanityChecks}.

Due to their flexibility and the plurality of data types on which they can be applied, Linear Interpretable Model-agnostic Explainer (LIME) \cite{trust} and SHapley Additive exPlanations (SHAP) \cite{unified} are the most frequently used interpretability methods in the industry \cite{XAI}. LIME offers to explain a prediction locally using models that are interpretable by design such as sparse regressions. The algorithm artificially generates data points in a neighborhood around the instance to explain and fits an interpretable model on these new examples. The SHAP method is inspired by the Shapley values\cite{shapley1953value} from economics and game theory. It aims to distribute fairly the rewards from a set of games to all the players. By associating the features of a model to the players to whom the gains are distributed, the prediction associated with an instance is decomposed by feature, which allows to compute each feature contribution.

\subsection{Related work}
The interpretability of the attention coefficients is still an open question \cite{notnotXP, notXP, Vashishth2019AttentionIA, serrano-smith-2019-attention}. Several methods based on self-attention coefficients have been proposed to explain the predictions made by Transformer-type models, such as attention flow and attention roll-out~\cite{flow_att}. These methods are based on complex aggregators to synthesize the information contained in the attention layers. Visualization tools allowing to dive in detail into the self-attention coefficients have been developed as well~\cite{vig-2019-multiscale}. Visualizing attention is the basis of saliency map approaches specific to Computer Vision for Vision Transformers~\cite{bastings-filippova-2020-elephant, DBLP:journals/corr/abs-2012-09838}. If these approaches allow to assign a local importance to input data, the quality of the explanation produced is not rigorously assessed. This raises the question of the evaluation of interpretability.

One way to assess the interpretability of a given method is to compare it quantitatively \cite{notXP} to common interpretable approaches such as SHAP. A given method would thus be considered interpretable if it strongly correlates with one target local feature importance method. This comparison criterion cannot be applied to approaches based on attention coefficients though, since attention is a \textit{positive} probability while LIME and SHAP contributions can be both \textit{positive} and \textit{negative}. Moreover, this is restrictive because the latter methods would represent ground truths to replicate. 

Human-grounded evaluation is needed in order to reach a rigorous science of interpretable machine learning and thus experimental approaches can be alternatives~\cite{been_kim_rigourous}. Therefore, interpretability methods can also be compared by evaluating to what extend they improve human performance during a specific annotation task \cite{ poursabzi2021manipulating}. In the context of NLP, it consists in asking humans to guess which class a text belongs to, in which words have been highlighted according to their local feature importance\cite{Quantifying, empirical_xai}. The response time and the average accuracy are then measured and the results are compared between the different methods. The method with the best response time/accuracy compromise is then considered as the most appropriate. This kind of experimental protocol has the advantage of quantifying the quality of an explanation, as long as the explanation is intended as a decision (in this case annotation) support tool. However, in the absence of robust statistical analysis, we find the work conducted by the authors to be insufficient to establish the validity of their conclusion.

\section{Methodology}
\label{sec:method}
We present a novel method, CLS-A, based on Transformer model attention coefficients. Then we define an experimental protocol inspired by the one previously introduced\cite{Quantifying} that we apply on three different annotation tasks of binary classification. Thereafter we provide implementation details of our experiment. Finally we introduce the evaluation protocol with data description, linear and non linear modeling.
\subsection{CLS-A}
We introduce our local feature importance method based on BERT self-attention that we call CLS-A. Since all the information needed for the classification task of a BERT goes through the \textit{CLS} token, we use the rich information contained in its related attention coefficients. We call context the distribution of attention linking a word in the input text to the rest of the sequence. Thus, CLS-A is constructed to represent the average context of the \textit{CLS} token.

Motivated by the fact that intermediate representations of deep neural networks become more abstract with network depth \cite{DBLP:journals/corr/YosinskiCNFL15}, we focus on the last layer of the BERT architecture. Since this last layer contains $h$ different self-attention matrices, the information has to be aggregated in order to build a one-dimensional local feature importance explanation. We aggregate the information contained by the $h$ attention heads by applying the average operator. This results in an average context of the classification token in the last layer of BERT. A weight is assigned to each word of the input text, representing its importance in the context that induced the prediction of the classifier.

Since the CLS token plays a central role in the computation of CLS-A, it is recommended that the BERT forward pass passes through the CLS token in order to perform its prediction. In the case where the BERT forward pass does not pass exclusively through the CLS token, an alternative would be to compute the average of all the coefficients of the attention heads.

\subsection{Experimental protocol}
The experimental protocol consists in asking participants to annotate one hundred texts in a binary classification task. Each text has some of its words colored with a more or less intense shade of blue, based on an underlying interpretability method or a random generator. The higher the coefficient of the method, the more strongly the word will be highlighted in blue. Accuracy and response time are measured to evaluate each method's ability to assist the participant in the annotation task. The higher the accuracy and the shorter the response time, the more relevant the method as it facilitates the human semantic processing of the text.

\paragraph{Setup and instructions.} 

\begin{figure}[t] 
    \centering
    \includegraphics[width=0.7\textwidth]{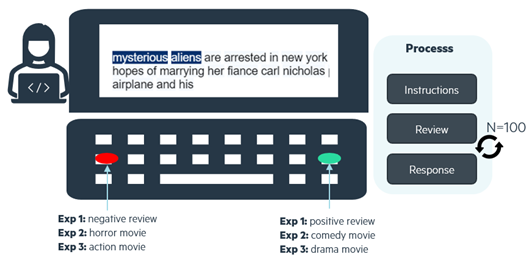}
    \caption{Scheme of the experimental protocol. Each participant labels 100 different texts after reading the instructions. The participant has two possible answers, depending on the  experiment he/she is participating in. The text is colored according to the interpretability method used to explain the classifier's prediction. The selected texts are all classified properly by the classifier.}
    \label{fig:schemaXP}
\end{figure}

All participants take part in the experiment in the same room and can be up to three at the same time. They are isolated in order to limit any other exogenous influence (visual, sound) and are placed in front of a computer as depicted in Figure~\ref{fig:schemaXP}. An explanation of the protocol is displayed on the screen to put the participants in the right conditions. In order to perform the annotation task, participants are asked to press either one of two buttons corresponding to the two possible answers as shown in Figure~\ref{fig:schemaXP}. The buttons correspond to keys on the keyboard of the computer used. Two colored stickers are stuck on the keys to help locate them. When a text is annotated (response given/key pressed), the next one is displayed on the screen.  

Three different classification tasks are tested. The first one (Experiment 1) is to evaluate the sentiment associated with a movie review. The participant must choose between the "positive" and "negative" sentiment. The second and third (Experiment 2 and 3) are film genre evaluations. In Experiment 2, the task is to distinguish between horror and comedy films, in Experiment 3 between action and drama. The information given in the explanation of the protocol differs depending on the classification task. Participants annotating film genres are asked to strive for a response time/accuracy trade-off. They are told that displayed colors can potentially be useful in the annotation task. Participants annotating movie review sentiments have no information about the colors displayed and no incentive to respond quickly.

\paragraph{Experiment characteristics.} All the 100 participants work in the same data science consulting firm. The first experiment involved 50 participants while the other two had 25 each. The participants were predominantly male: about 76\% compared to 24\% for women. They were between 22 and 40  years old and none of the subjects were cognitively impaired to our knowledge.

Every participant is asked to annotate 100 different texts in a binary classification task. The asked classification task remains the same during the whole experiment. A participant cannot see the same text twice during the experiment. Each text has its words colored in different shades of blue. This coloring is proportional to the coefficients chosen at random among a random baseline, SHAP, LIME and CLS-A  built on the DistilBERT classifier attention. The random baseline assigns randomly a coefficient to each word. The feature importance obtained from LIME and SHAP to color the text are truncated to be exclusively positive. Thus, the negative SHAP and LIME values are set to 0 in order to highlight only the words contributing positivly to the predicted class. This truncation of SHAP and LIME make them more comparable to the attention coefficients. Participants are thus subjected to exogenous attentional orientation effects in order to compare the methods one-to-one. An example of the text displayed during the experiment is plotted in Figure~\ref{fig:schemaXP}.
 
The classes of the various classification tasks are all equally represented among the displayed texts. Only the instances that the models predicted correctly were selected. We wanted to look at the effect of the review length and the prediction probabilities in Experiments 2 and 3. The reviews corresponding to the sentiment analysis task contain between 32 and 50 words. The text sequence lengths of the other classification tasks vary between 19 and 145 words. The probability scores of belonging to the target class are highly polarized for the sentiment analysis and the horror/comedy classification while probability score is more uniformly distributed for the action/drama classification task.
We assume that an interpretability method provides good explanations to the extent that it helps an annotator to go faster and be more efficient. An explanation will then be the object of a semantic congruence between the label to be predicted and the words highlighted. Therefore, the response time is precisely measured for each text and the correctness of the answers is assessed.

\subsection{Implementation details}
The three classifiers analyzed during the three annotation tasks were based on a DistilBERT\cite{sanh_distilbert_2020}, a stack of 6 encoders being the result of a distillation of a classical BERT. Each pre-trained DistilBERT was retrieved from Hugging Face\footnote{www.huggingface.co/}. A dense layer was added to perform the classification and fine-tune each model. The forward pass was defined as getting the embedding of the CLS token to perform the classification task. The library used to compile and fine-tune the models were Keras on the TensorFlow framework. Each model was trained with an initial learning rate of $10^{-5}$ and a reducing learning strategy when reaching a plateau. The number of epochs was for each model of 5 and the batch size was 32. The models were trained with a categorical crossentropy loss and the Adam optimizer. The first model was fine-tuned for sentiment analysis on the IMDB database\cite{maas_learning_2011}. The second and third one were fine-tuned to perform movie genre classification on a Kaggle dataset\footnote{www.kaggle.com/competitions/movie-genre-classification/overview}.

For each text, SHAP was computed with the \verb|shap| library~\cite{unified}. The Shapley values were computed in a permutation way. Finally, LIME was computed with the \verb|lime|~\cite{trust} library. The whole experiment was performed on the \verb|psychopy|~\cite{peirce2019psychopy2} framework on Python.

\subsection{Data analyses}
Here we define the methods used to analyze the data produced by the experimental protocol presented above. Each experiment produces $n \times 100$ answers, with $n$ the number of participants, and 100 the number of plotted text samples during each experiment. The indicators of interest are the labeling time, which we call "reaction time", and whether or not the participant is wrong, which we call "accuracy". These variables of interest are then analyzed through their relationship with other characteristics such as features about the text (sequence length, probabiltiy score, trial number, relative position of impacting word) and the interpretability method used to color it.

\paragraph{Data description.} The descriptive analysis is first performed by calculating the average reaction time and the average accuracy. The one-tail $t$-test is then used to compare the distributions of reaction times between interpretability methods and the random baseline in order to have statistically significant comparisons. The one-tailed $t$-test is a statistical hypothesis test used to determine whether the mean of a first sample is lower the mean of a second one. This test is applied here to the average difference between each interpretability method and the random baseline, per participant, per experiment.
\paragraph{Linear modeling.} The impact of interpretability methods on reaction time is estimated with a linear regression by incorporating the effect of other explanatory variables. 



The random baseline is used for reference. Thus, the coefficients of the linear regression associated with the method used to color the text are expressed with respect to this baseline. 

For each experiment, a linear model is built per participant to explain the reaction time to the labeling task. The explanatory variables of the models for an experiment are the same for all participants. The mean value of the regression parameters and their distribution are then analyzed using the one-tail $t$-test presented above.
\paragraph{Non-linear modeling.} Decision tree boosting algorithms allow to model complex and non-linear phenomena. We apply this type of algorithm to model the participant accuracy. This binary classification problem is addressed via Explainable Boosting Machine (EBM) \cite{nori2019interpretml}. EBM obtain performance levels equivalent to other boosting approaches based on decision trees, while decomposing its prediction into contributions of the explanatory variables. EBM is a generalized additive model (GAM) of the form:
\begin{equation}
g(\mathbb{E}[y]) = \beta_0 + \sum_{i}f_{i}(x_i) + \sum_{i,j, i\ne j}f_{i,j}(x_{i},x_{j})
\end{equation}

Where:
\begin{itemize}
    \item $y$ is the variable indicating whether a participant has successfully completed its labeling 
    \item $g$ is the link function
    \item $\beta_0$ is the intercept
    \item $f_i$ is the feature response function of the variable $x_i,$
    \item $f_{i,j}$ is the pairwise interaction function of the two variables $x_{i}$ and $x_{j}$
\end{itemize}
 A response curve reflects the impact of a given explanatory variable by plotting the evolution of its contribution to the target variable. One model is fitted per method and per experiment in order to compare the response curves of the methods within a given experiment. Each model has to be trained with the same explanatory variables. Since the participants generally perform their annotation tasks accurately, the data are largely imbalanced. Sub-sampling is then performed to run the EBM on a balanced dataset with as many right and wrong answers. Since this sub-sampling induces sampling bias, this operation is run again 50 times. Average response curves and standard deviations are then calculated to produce the results in the analysis. 

\section{Results}
\label{sec:results}
This section reports the analysis of the data produced by the three experiments with the data analysis methods introduced in Section~\ref{sec:method}. LIME and SHAP are compared to CLS-A and a random baseline. We show that CLS-A improves both speed and accuracy of annotation in a statistically significant way compared to the random baseline. CLS-A, SHAP and LIME result in statistically similar response times and accuracy. Furthermore, we highlight the relationship between the quality of an explanation and the certainty of the classifier's prediction.

\subsection{Data description}
\begin{table}[t]
\centering
\begin{tabular}{@{}cccccc@{}}
\toprule
Metrics                            & Experiment & CLS-A          & LIME          & SHAP          & Random \\ \midrule
\multirow{3}{*}{Reaction Time (s)} & Exp 1      & \textbf{10.52} & 10.83         & 10.74         & 11.21  \\
                                   & Exp 2      & 9.13           & 8.72          & 8.74          & 8.79   \\
                                   & Exp 3      & \textbf{10.85} & 11.25         & 11.66         & 12.28  \\ \midrule
\multirow{3}{*}{Accuracy (\%)}     & Exp 1      & \textbf{96.1}  & \textbf{96.4} & 95.4          & 95.2   \\
                                   & Exp 2      & \textbf{80.9}  & 79.1          & \textbf{80.4} & 79.9   \\
                                   & Exp 3      & \textbf{85.9}  & \textbf{84.9} & 80.0          & 83.0   \\ \bottomrule
\end{tabular}
\caption{Average reaction time and accuracy per experiment per method.}
      \label{tab:table1}
\end{table}

    
This section provides an exploratory analysis of participant's responses to the three experiments as presented in Section~\ref{sec:method}. The first experiment consisted of responses from 100 participants while the other two had 50 each. Table~\ref{tab:table1} relates the average reaction time and the average accuracy per experiment and per method used to color the text. This shows that the average reaction time related to CLS-A is lower for experiments 1 and 3. Accuracy is also on average higher for participants who were exposed to CLS-A. The random baseline induces less accurate and slower responses overall.

\begin{figure}[t]
\centerline{\includegraphics[width=1\linewidth, height=7cm]{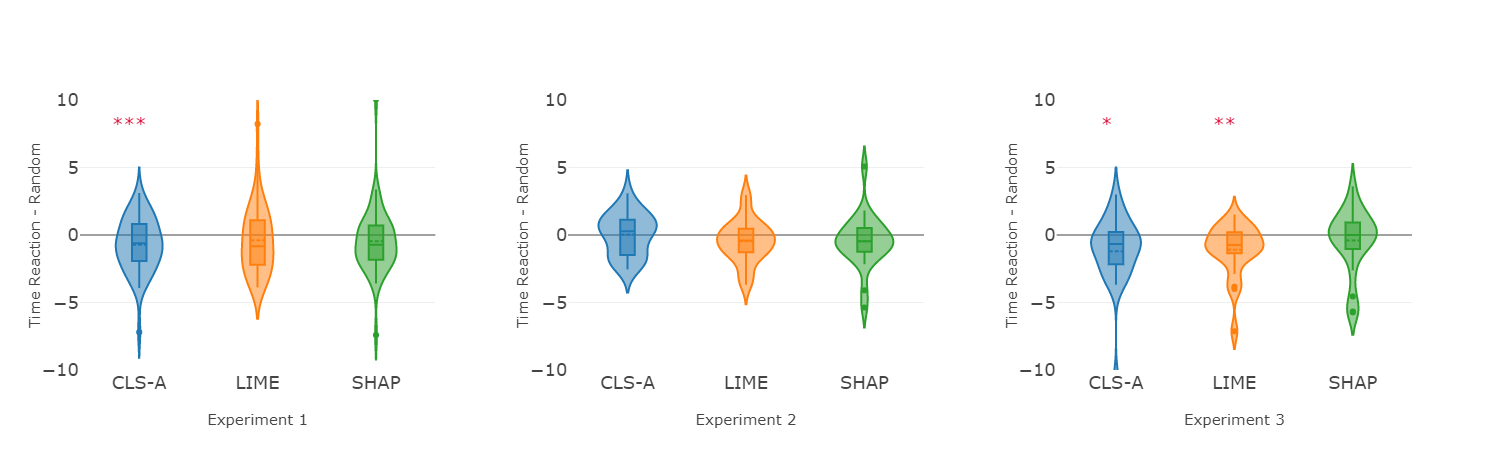}}
\caption{Distribution of mean reaction time deviation from random baseline by participant, by experiment. The results of the one-tailed $t$-test are represented with stars above the violin plots. With $p$ as the $p$-value of the $t$-test, *$p<5$\%, **$p<1$\%, and ***$p<0.5$\%}
\label{fig:distrib time reaction minus random}
\end{figure}

We also compare the distributions of the average response time of the CLS-A, LIME and SHAP methods in comparison to the random baseline. We perform this distribution comparison using the one-tailed $t$-test on the average difference between the coloring method and the baseline, per participant as presented in Section~\ref{sec:method} . Figure~\ref{fig:distrib time reaction minus random} plots the distribution of the average reaction time deviation from the random baseline with the results of the $t$-tests, by method and by experiment. 


The mean difference in reaction time between CLS-A and the random baseline is statistically significant in the first experiment with a risk level of 0.1\%. The second experiment shows no significant difference in response time between the different methods and the random baseline. Finally, the third experiment highlights a mean distribution difference significantly lower than 0 between CLS-A, SHAP and the random baseline at risk levels of 5\% and 1\% respectively. Therefore, participants went faster on average in the text annotation task in Experiments 1 and 3 when they were exposed to CLS-A compared to the baseline. This difference from baseline is exclusive to CLS-A for Experiment 1, and shared with LIME for Experiment 3.
    
 
This section presents the analysis of the difference in reaction time between the interpretability methods and the random baseline by linear modeling . We also assess the impact of other important features affecting the reaction time, such as review length and probability score of the target label.

For each experiment, we run a linear regression per participant as presented in Section~\ref{sec:method} to model the reaction time to the labeling task. The explanatory variables differ only slightly from one experiment to another. The explanatory variables are information about the text and the interpretability method used to color it and are presented in Table~\ref{tab:variables} in Appendix~\ref{sec:appendix}. 

\begin{figure}[t]
\centerline{\includegraphics[width=1\linewidth, height=8cm]{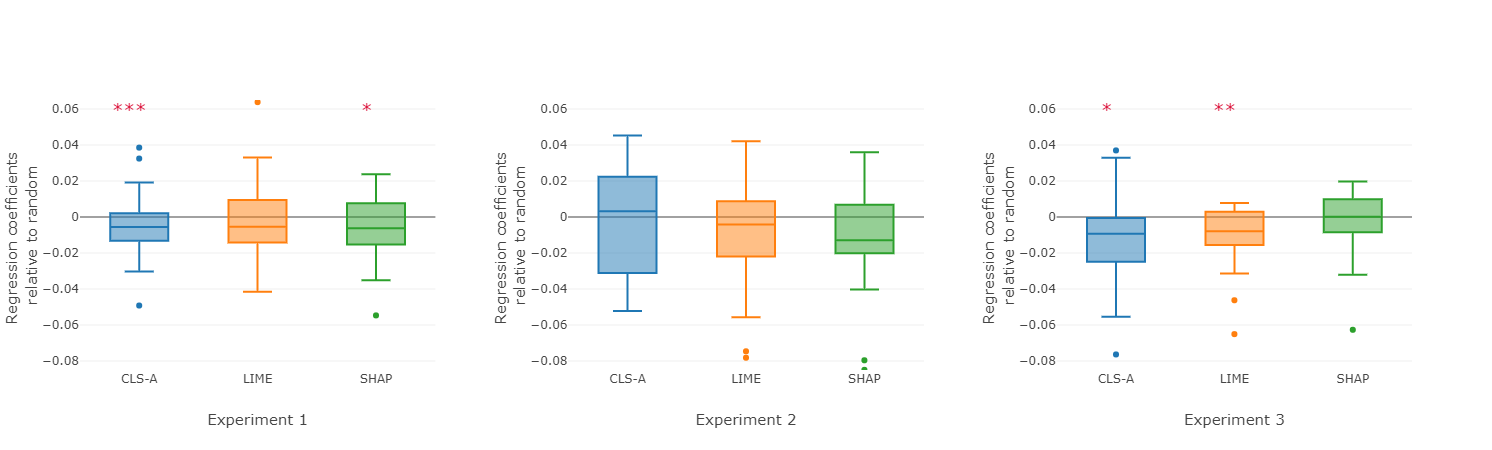}}
\caption{Distribution of linear modeling on reaction times coefficients of each interpretability method variable with respect to the baseline. The results of the one-tailed $t$-test are represented with stars above the box plots. With $p$ as the $p$-value of the one-tailed $t$-test, *$p<5$\%, **$p<1$\%, and ***$p<0.5$\%}
\label{fig:all_reg_methods}
\end{figure}

\paragraph{Method effect.} The performance metrics of the linear regressions on reaction times are presented in Table~\ref{tab:table_methods_R2} Appendix~\ref{sec:appendix} and the distributions of the average reaction time deviation from the random baseline per experiment are shown in Table~\ref{fig:distrib time reaction minus random}. The average and median coefficients of all methods, over all experiments are negative. Since all coefficients are calculated with respect to the random baseline, this suggests a lower average response time when participants are exposed to interpretability methods. CLS-A has the greatest impact compared to the random baseline for the first and third experiments, while SHAP induces faster responses for the second experiment. In order to evaluate the statistical significance of the results obtained, we analyze the distribution of the parameters of each linear model for each experiment. We then assess the mean sign of these distributions by performing a one-tailed $t$-test in the same way as in the previous section. The Figure~\ref{fig:all_reg_methods} shows the distributions of coefficients of linear regressions on reaction times associated with the interpretability method used to color the text.

\begin{figure}[t]
\centerline{\includegraphics[width=1\linewidth, height=8cm]{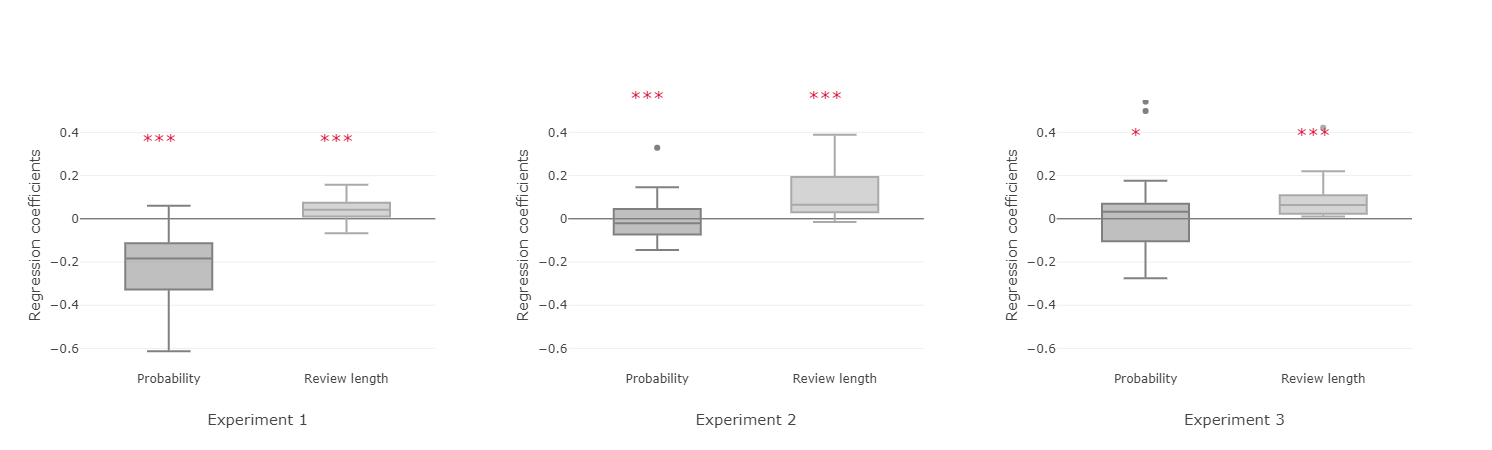}}
\caption{Distribution of probability score and review length coefficients in linear modeling. The results of the one-tailed $t$-test are represented with stars above the box plots. Noting $p$ as the $p$-value of the statistical test, the notations are as follows. *$p<5$\%, **$p<1$\%, and ***$p<0.5$\%}
\label{fig:all_exp_methods_reg_proba_rev}
\end{figure}

The average linear regression coefficients on reaction times of CLS-A, SHAP and LIME are significantly negative for Experiment 1 and 3. The level of associated risk is however lower for CLS-A, with 0.1\% against 1\% and 5\% respectively for LIME and SHAP. The results for the CLS-A method are broadly consistent with the previous exploratory analysis. Participants took therefore less time on average to complete their annotation task on Experiment 1 and 3 when important words in the text were colored via the CLS-A method compared to the random baseline. However, the results differ for SHAP and LIME, and Experiment 2 does not show a statistically significant difference between these 3 methods compared to the random baseline.
\paragraph{Probability score and review length effects.} We similarly assess the distributions of two more features in the linear regression model on reaction times, namely the probability of belonging to the target class, and the length of the word sequence (review). The Figure~\ref{fig:all_exp_methods_reg_proba_rev} represents the distribution of these coefficients, by experiment. The significance of the means of the distributions is assessed with one-tailed paired $t$-test in the same way than CLS-A, LIME and SHAP.


The sign of each of the two coefficients is consistent across all three experiments. The effect of the probability score variable on response time is negative on average. This impact is statistically significant for experiment 1 and 2. The effect of the sequence length variable is positive and statistically significant on all experiments. Thus, all things being equal, the higher the probability score of belonging to the target class, the lower the reaction time. This highlights the relationship between the quality of an explanation and the certainty of a prediction from a time reaction perspective. In the same way, all things being equal, the annotation time increases with the length of the textual sequence processed.

Therefore, the linear modeling allowed to highlight that CLS-A fosters quicker responses on average compared to the random baseline. This result is found in experiment 1 and 3. If LIME and SHAP generate faster responses on average in these two experiments, CLS-A does so as well at a lower level of statistical risk. Furthermore, the probability score of the target class impacts negatively the reaction time.

\subsection{Non-linear modeling}
This section compares CLS-A and the random baseline through the prism of the participant's accuracy, which is modelled using non-linear Explainable Boosting Machines (EBM) introduced in Section~\ref{sec:method}. 
The explanatory variables used to explain participant's response to the experiment are presented in Appendix~\ref{sec:appendix}. We fit one model per method and per experiment in order to compare the response curves of the methods within a given experiment. Each model is trained with the same explanatory variables. The same analysis integrating LIME and SHAP is in Appendix~\ref{sec:appendix}.   

\begin{figure}[t!]
\centerline{\includegraphics[width=1\linewidth, height=5cm]{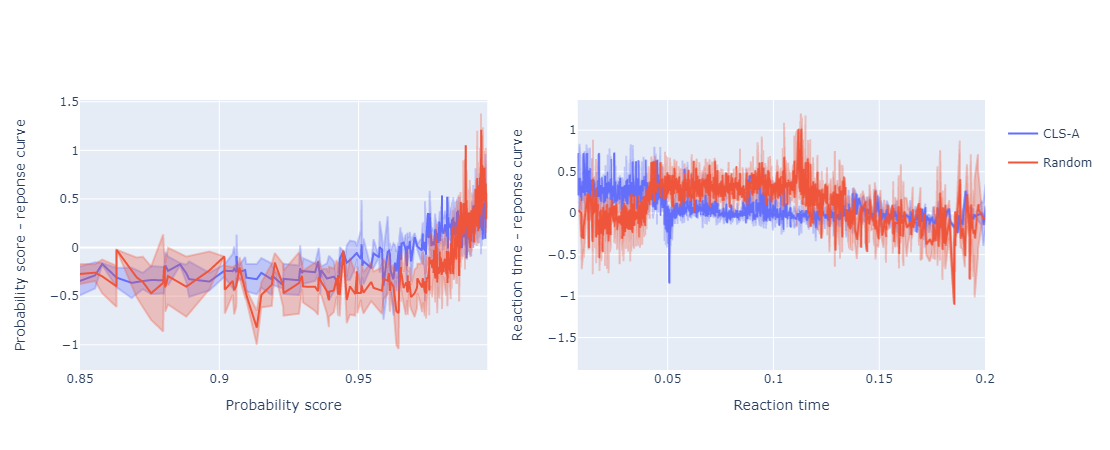}}
\caption{Explainable boosting machine response curves of probability score and time reaction in the first experiment. The contribution of the probability score variable becomes positive at a lower probability threshold for CLS-A compared to the random generator. The contributions of the reaction time variables are positive for the fast reactions for the CLS-A method unlike the random generator.}
\label{fig:response_curve_xp1}
\centerline{\includegraphics[width=1\linewidth, height=5cm]{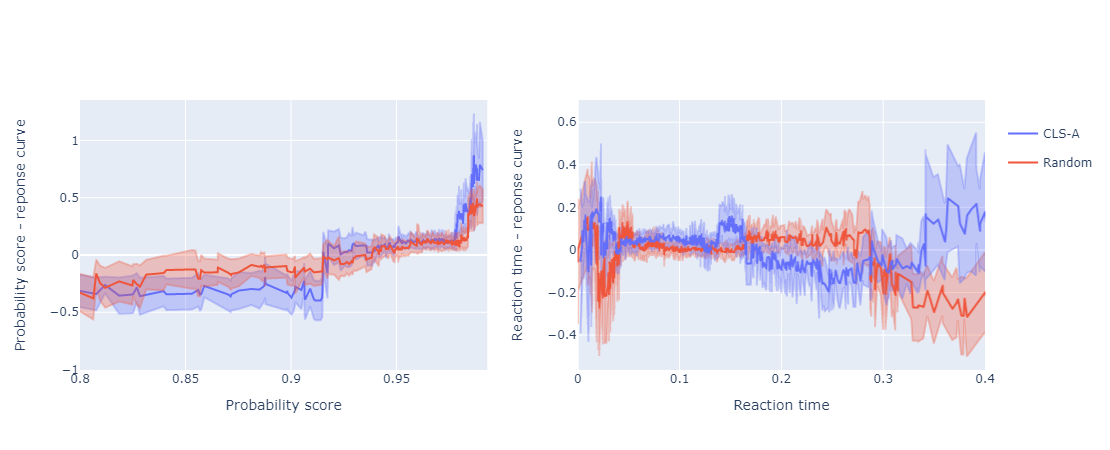}}
\caption{Explainable boosting machine response curves of probability score and time reaction in the second experiment. The contribution of the probability score increases at a faster rate than the random generator. CLS-A favors more fast reactions.}
\label{fig:response_curve_xp2}
\centerline{\includegraphics[width=1\linewidth, height=5cm]{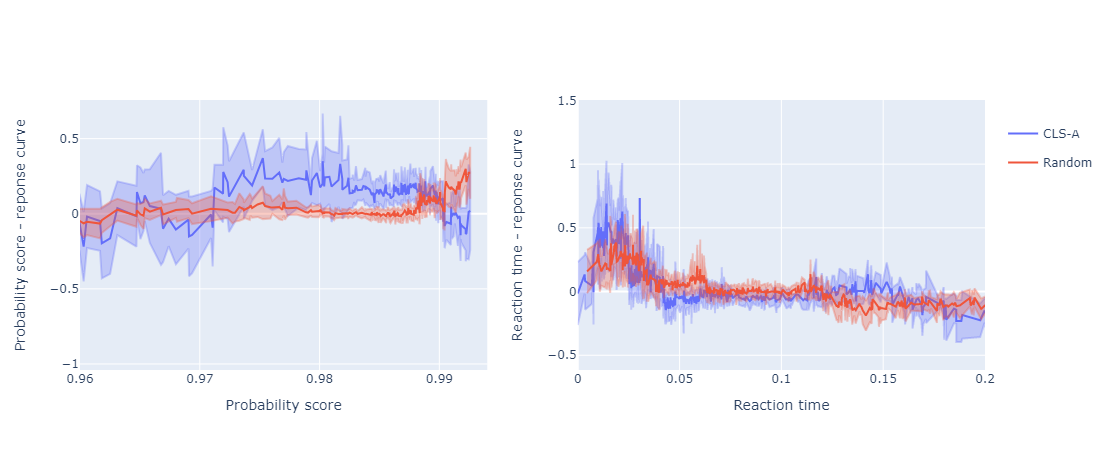}}
\caption{Explainable boosting machine response curves of probability score and time reaction in the third experiment. The contribution of the probability score variable increases at a faster rate than the random generator and decreases for very high prediction probability score unlike the random generator. The contributions are slightly higher for fast response times for CLS-A.}
\label{fig:response_curve_xp3}
\end{figure}

We focus on this section on the impact of the probability score and the reaction time on the annotation task. Figure~\ref{fig:response_curve_xp1}, \ref{fig:response_curve_xp2} and \ref{fig:response_curve_xp3} represent the EBM response curves of the  probability score and the reaction time, by method, by experiment. These curves represent the contributions to the probability scores that the participant performs the annotation task well. The response curves of the variables sequence length, relative position of important words and trial number are shown in Appendix~\ref{sec:appendix}.The interval around the mean curve integrates the standard deviation measured on 50 sampling iterations for a given model. For each of the analyzed variables, we focus on the comparison between CLS-A and the random baseline. The analyses of LIME and SHAP are in  Appendix~\ref{sec:appendix}. 



\paragraph{Sentiment analysis.} The first experiment highlights a higher target class probability score contribution for CLS-A compared to the random baseline in Figure~\ref{fig:response_curve_xp1}. The response curves tend to merge for the polarized probability scores and CLS-A falls a bit below the baseline for the probability score distribution tail. Accuracy contributions related to reaction time under CLS-A influence are higher for fast reactions, and lower for medium time reactions. Accuracy contribution tend to the same for very long response times. We can conclude that the interest of CLS-A compared to the random generator lies in the relatively low probabilities in the first experiment. Note however that the probability score distribution is very high in the first experiment, and covers very few non-polarized predictions. Moreover, the contribution of CLS-A is significant for fast predictions, and tends to vanish gradually.



\paragraph{Movie genre classification, action vs drama.} The second experiment has a more dispersed distribution of target class probability scores than the first experiment. The CLS-A response curve associated with the target class probability score variable is higher for polarized predictions as shown in in Figure~\ref{fig:response_curve_xp2}. The contribution of CLS-A compared to the baseline thus seems to be related to the certainty of the classifier prediction. The area in which the CLS-A response curve is higher corresponds to the majority of the probability score distribution of the target variable. Finally, the accuracy contributions of the reaction time variables are higher for CLS-A for short and very long responses. Finally and similarly to the first experiment, CLS-A has a strong impact to form rapid responses when labeling a high target class probability score text.



\paragraph{Movie genre classification, horror vs comedy.} The distribution of the target class probability score variable is less dispersed in the last experiment than in the second one. Figure~\ref{fig:response_curve_xp3} depicts that the response curve is higher for CLS-A for high probability scores and falls below it at the distribution tail in a slightly more marked way than the first experiment. Finally, the effect of CLS-A on the response time variable with respect to the baseline in experiment 3 is relatively similar to Experiment 2. Short and very long answers are more accurate with CLS-A compared to the random baseline. 

Altogether, the analysis of the response curves highlights the non-linear relationships between the explanatory variables and the target variable. The interest of CLS-A seems strongest for high probability scores of classification, but tends to decrease for texts whose probability score is at the tail of the distribution. In the same way as in the linear modeling analysis, the interest of CLS-A seems to be less important or even non-existent for texts whose probability scores are relatively low. This highlights a strong link between the quality of an explanation and the certainty of a prediction, insofar as participants respond less well overall when exposed to such texts colored by the CLS-A method. The fact that the interest of CLS-A fades for high probabilities for experiment 1 and 3 may be related to the intrinsic ease of classifying text with such a high probability score. We also found that the impact of the CLS-A method is concentrated around fast and slow responses, and tends to merge with the baseline for medium fast responses, confirming the facilitating effect of the method on semantic processing. 

\section{Conclusion and future work}
We applied an experimental protocol to compare a new method called CLS-A based on Transformer self-attention to SHAP, LIME and a random baseline. We found that CLS-A helps in the same proportions as SHAP and LIME to annotate text on three different tasks and is significantly better than a random baseline. This work adds to the literature aiming to evaluate the interpretability of attention coefficients in recent deep learning models. Moreover, analyses have shown that the interest of an CLS-A relies highly on the probability score of the prediction. The higher the probability score, the more relevant the explanation. As far as we know, this is the first time that the relationship between the quality of an explanation and the certainty of its associated prediction has come to light. 

There is currently no consensus on a quantitative method to validate interpretability techniques. We believe human-based experiments measuring the similarity of processing between the model and human reasoning is the best option available. The present study confirms its relevance, especially when combined with statistical analysis.

This work has been done by analyzing binary classifiers. Analyzing multi-class classifiers could likewise lead to more refined conclusions. Finally, it would be insightful to include in such experiments other feature importance methods usually used in NLP based, for example, on gradient computation in order to to benchmark them in relation to common approaches. To our knowledge, there is no experimental protocol to compare other type of explainability method such as counterfactual explanations. We plan in a future work to build such a protocol, which will also allow to assess the quality of a given counterfactual explanation.

\section{Acknowledgments}
We thank Gabriel Olympie and Jean-Baptiste Gette for their advice and support. We also thank all the participants in the experiment who made this work possible.

\section{Ethic statement}
Each participant signed a consent form containing an overview of the project and the intended use of the data they would generate. The data was anonymized and processed only by our team. The data produced is stored in a file in respect with the GDPR regulations in force. Participation in the study was fully voluntary. It was possible to stop performing the labeling tasks at any time.  

\bibliography{template}

\begin{thebibliography}{10}

\bibitem{poursabzi2021manipulating}
Forough Poursabzi-Sangdeh, Daniel~G Goldstein, Jake~M Hofman, Jennifer~Wortman
  Wortman~Vaughan, and Hanna Wallach.
\newblock Manipulating and measuring model interpretability.
\newblock In {\em Proceedings of the 2021 CHI conference on human factors in
  computing systems}, pages 1--52, 2021.

\bibitem{Attention}
Ashish Vaswani, Noam Shazeer, Niki Parmar, Jakob Uszkoreit, Llion Jones,
  Aidan~N Gomez, \L~ukasz Kaiser, and Illia Polosukhin.
\newblock Attention is all you need.
\newblock In I.~Guyon, U.~Von Luxburg, S.~Bengio, H.~Wallach, R.~Fergus,
  S.~Vishwanathan, and R.~Garnett, editors, {\em Advances in Neural Information
  Processing Systems}, volume~30. Curran Associates, Inc., 2017.

\bibitem{Neural}
Dzmitry Bahdanau, Kyunghyun Cho, and Yoshua Bengio.
\newblock Neural machine translation by jointly learning to align and
  translate.
\newblock {\em CoRR}, abs/1409.0473, 2014.

\bibitem{notnotXP}
Sarah Wiegreffe and Yuval Pinter.
\newblock Attention is not not explanation.
\newblock In {\em Proceedings of the 2019 Conference on Empirical Methods in
  Natural Language Processing and the 9th International Joint Conference on
  Natural Language Processing (EMNLP-IJCNLP)}, pages 11--20, Hong Kong, China,
  November 2019. Association for Computational Linguistics.

\bibitem{notXP}
Sarthak Jain and Byron~C. Wallace.
\newblock Attention is not explanation.
\newblock In {\em North American Chapter of the Association for Computational
  Linguistics}, 2019.

\bibitem{DistilBERT}
V.~Sanh, L.~Debut, J.~Chaumond, and T.~Wolf.
\newblock Distilbert, a distilled version of bert: smaller, faster, cheaper and
  lighter.
\newblock 2019.

\bibitem{devlin_bert_2019}
Jacob Devlin, Ming-Wei Chang, Kenton Lee, and Kristina Toutanova.
\newblock {BERT}: {Pre}-training of {Deep} {Bidirectional} {Transformers} for
  {Language} {Understanding}.
\newblock Technical Report arXiv:1810.04805, arXiv, May 2019.
\newblock arXiv:1810.04805 [cs] type: article.

\bibitem{wang-etal-2018-glue}
Alex Wang, Amanpreet Singh, Julian Michael, Felix Hill, Omer Levy, and Samuel
  Bowman.
\newblock {GLUE}: A multi-task benchmark and analysis platform for natural
  language understanding.
\newblock In {\em Proceedings of the 2018 {EMNLP} Workshop {B}lackbox{NLP}:
  Analyzing and Interpreting Neural Networks for {NLP}}, pages 353--355,
  Brussels, Belgium, November 2018. Association for Computational Linguistics.

\bibitem{deep_mind_transfo}
Mary Phuong and Marcus Hutter.
\newblock Formal algorithms for transformers.
\newblock 2022.

\bibitem{guidotti_counterfactual_2022}
Riccardo Guidotti.
\newblock Counterfactual explanations and how to find them: literature review
  and benchmarking.
\newblock {\em Data Mining and Knowledge Discovery}, April 2022.

\bibitem{molnar_interpretable_2020}
Christoph Molnar.
\newblock {\em Interpretable {Machine} {Learning}}.
\newblock Lulu.com, 2020.
\newblock Google-Books-ID: jBm3DwAAQBAJ.

\bibitem{MLint_Survey}
Diogo~V. Carvalho, Eduardo~M. Pereira, and Jaime~S. Cardoso.
\newblock Machine learning interpretability: A survey on methods and metrics.
\newblock {\em Electronics}, 8(8), 2019.

\bibitem{Axio}
Mukund Sundararajan, Ankur Taly, and Qiqi Yan.
\newblock Axiomatic attribution for deep networks.
\newblock In Doina Precup and Yee~Whye Teh, editors, {\em Proceedings of the
  34th International Conference on Machine Learning}, volume~70 of {\em
  Proceedings of Machine Learning Research}, pages 3319--3328. PMLR, 06--11 Aug
  2017.

\bibitem{Importance}
Avanti Shrikumar, Peyton Greenside, and Anshul Kundaje.
\newblock Learning important features through propagating activation
  differences.
\newblock In Doina Precup and Yee~Whye Teh, editors, {\em Proceedings of the
  34th International Conference on Machine Learning}, volume~70 of {\em
  Proceedings of Machine Learning Research}, pages 3145--3153. PMLR, 06--11 Aug
  2017.

\bibitem{SanityChecks}
Julius Adebayo, Justin Gilmer, Michael Muelly, Ian Goodfellow, Moritz Hardt,
  and Been Kim.
\newblock Sanity checks for saliency maps.
\newblock In S.~Bengio, H.~Wallach, H.~Larochelle, K.~Grauman, N.~Cesa-Bianchi,
  and R.~Garnett, editors, {\em Advances in Neural Information Processing
  Systems}, volume~31. Curran Associates, Inc., 2018.

\bibitem{trust}
Marco~Tulio Ribeiro, Sameer Singh, and Carlos Guestrin.
\newblock "why should i trust you?": Explaining the predictions of any
  classifier.
\newblock In {\em Proceedings of the 22nd ACM SIGKDD International Conference
  on Knowledge Discovery and Data Mining}, KDD '16, page 1135–1144, New York,
  NY, USA, 2016. Association for Computing Machinery.

\bibitem{unified}
Scott~M Lundberg and Su-In Lee.
\newblock A unified approach to interpreting model predictions.
\newblock In I.~Guyon, U.~Von Luxburg, S.~Bengio, H.~Wallach, R.~Fergus,
  S.~Vishwanathan, and R.~Garnett, editors, {\em Advances in Neural Information
  Processing Systems}, volume~30. Curran Associates, Inc., 2017.

\bibitem{XAI}
Umang Bhatt, Alice Xiang, Shubham Sharma, Adrian Weller, Ankur Taly, Yunhan
  Jia, Joydeep Ghosh, Ruchir Puri, Jos\'{e} M.~F. Moura, and Peter Eckersley.
\newblock Explainable machine learning in deployment.
\newblock In {\em Proceedings of the 2020 Conference on Fairness,
  Accountability, and Transparency}, FAT* '20, page 648–657, New York, NY,
  USA, 2020. Association for Computing Machinery.

\bibitem{shapley1953value}
S~Shapley~Ll.
\newblock A value for n-person games.
\newblock {\em Contributions to the Theory of Games II, Annals of Mathematical
  Studies}, 28, 1953.

\bibitem{Vashishth2019AttentionIA}
Shikhar Vashishth, Shyam Upadhyay, Gaurav~Singh Tomar, and Manaal Faruqui.
\newblock Attention interpretability across nlp tasks.
\newblock {\em ArXiv}, abs/1909.11218, 2019.

\bibitem{serrano-smith-2019-attention}
Sofia Serrano and Noah~A. Smith.
\newblock Is attention interpretable?
\newblock In {\em Proceedings of the 57th Annual Meeting of the Association for
  Computational Linguistics}, pages 2931--2951, Florence, Italy, July 2019.
  Association for Computational Linguistics.

\bibitem{flow_att}
Samira Abnar and Willem~H. Zuidema.
\newblock Quantifying attention flow in transformers.
\newblock {\em CoRR}, abs/2005.00928, 2020.

\bibitem{vig-2019-multiscale}
Jesse Vig.
\newblock A multiscale visualization of attention in the transformer model.
\newblock In {\em Proceedings of the 57th Annual Meeting of the Association for
  Computational Linguistics: System Demonstrations}, pages 37--42, Florence,
  Italy, July 2019. Association for Computational Linguistics.

\bibitem{bastings-filippova-2020-elephant}
Jasmijn Bastings and Katja Filippova.
\newblock The elephant in the interpretability room: Why use attention as
  explanation when we have saliency methods?
\newblock In {\em Proceedings of the Third BlackboxNLP Workshop on Analyzing
  and Interpreting Neural Networks for NLP}, pages 149--155, Online, November
  2020. Association for Computational Linguistics.

\bibitem{DBLP:journals/corr/abs-2012-09838}
Hila Chefer, Shir Gur, and Lior Wolf.
\newblock Transformer interpretability beyond attention visualization.
\newblock {\em CoRR}, abs/2012.09838, 2020.

\bibitem{been_kim_rigourous}
Finale Doshi-Velez and Been Kim.
\newblock Towards a rigorous science of interpretable machine learning, 2017.

\bibitem{Quantifying}
P.~Schmidt and F.~Biessmann.
\newblock Quantifying interpretability and trust in machine learning systems.
\newblock 2019.

\bibitem{empirical_xai}
Andrew Bell, Ian Solano-Kamaiko, Oded Nov, and Julia Stoyanovich.
\newblock It’s just not that simple: An empirical study of the
  accuracy-explainability trade-off in machine learning for public policy.
\newblock In {\em 2022 ACM Conference on Fairness, Accountability, and
  Transparency}, FAccT '22, page 248–266, New York, NY, USA, 2022.
  Association for Computing Machinery.

\bibitem{DBLP:journals/corr/YosinskiCNFL15}
Jason Yosinski, Jeff Clune, Anh~Mai Nguyen, Thomas~J. Fuchs, and Hod Lipson.
\newblock Understanding neural networks through deep visualization.
\newblock {\em CoRR}, abs/1506.06579, 2015.

\bibitem{sanh_distilbert_2020}
Victor Sanh, Lysandre Debut, Julien Chaumond, and Thomas Wolf.
\newblock {DistilBERT}, a distilled version of {BERT}: smaller, faster, cheaper
  and lighter, February 2020.
\newblock arXiv:1910.01108 [cs].

\bibitem{maas_learning_2011}
Andrew~L. Maas, Raymond~E. Daly, Peter~T. Pham, Dan Huang, Andrew~Y. Ng, and
  Christopher Potts.
\newblock Learning {Word} {Vectors} for {Sentiment} {Analysis}.
\newblock In {\em Proceedings of the 49th {Annual} {Meeting} of the
  {Association} for {Computational} {Linguistics}: {Human} {Language}
  {Technologies}}, pages 142--150, Portland, Oregon, USA, June 2011.
  Association for Computational Linguistics.

\bibitem{peirce2019psychopy2}
Jonathan Peirce, Jeremy~R Gray, Sol Simpson, Michael MacAskill, Richard
  H{\"o}chenberger, Hiroyuki Sogo, Erik Kastman, and Jonas~Kristoffer
  Lindel{\o}v.
\newblock Psychopy2: Experiments in behavior made easy.
\newblock {\em Behavior research methods}, 51(1):195--203, 2019.

\bibitem{nori2019interpretml}
Harsha Nori, Samuel Jenkins, Paul Koch, and Rich Caruana.
\newblock Interpretml: A unified framework for machine learning
  interpretability.
\newblock {\em arXiv preprint arXiv:1909.09223}, 2019.

\end{thebibliography}
\bibliographystyle{unsrt}
\newpage
\appendix
\section{Appendix}
\label{sec:appendix}

\subsection{Regression results} \label{A}
\begin{table}[H]
\centering
\begin{tabular}{@{}ccc@{}}
\toprule
Metrics                            & Experiment & \textit{$R^2$} \\ \midrule
\multirow{3}{*}{Reaction Time (s)} & Exp 1      & 4.23e-1        \\
                                   & Exp 2      & 6.01e-1        \\
                                   & Exp 3      & 6.59e-1        \\ \bottomrule
\end{tabular}
\caption{R-square of linear regression explaining participant reaction time}
      \label{tab:table_methods_R2}
\end{table}


\begin{table}[H]
      \centering
      \begin{tabular}{lllllll}
        \toprule
        \cmidrule(){1-4}
        Metrics & Experiment 1 & Experiment 2 & Experiment 3\\
        \midrule
         Intercept & \hfil\num{3.68e-1} & \hfil\num{1.50e-1} & \hfil\num{9.83e-2}\\
         Method CLS-A  & \hfil\num{-6.10e-3} & \hfil\num{-2.70e-3} & \hfil\num{-1.07e-2}\\
         Method LIME & \hfil\num{-3.43e-3} & \hfil\num{-1.00e-2} & \hfil\num{-1.04e-2}\\
         Method SHAP & \hfil\num{-4.79e-3} & \hfil\num{-9.39e-3} & \hfil\num{-3.21e-3}\\
         Probability  & \hfil\num{-2.24e-1} & \hfil\num{4.89e-3} & \hfil\num{3.19e-2}\\
         Accurate answer & \hfil\num{-5.25e-2} & \hfil\num{-1.88e-2} & \hfil\num{-1.66e-2}\\
         Review length & \hfil\num{4.38e-2} & \hfil\num{1.15e-1} & \hfil\num{8.25e-2}\\
         First word position & \hfil\num{5.21e-3} & \hfil\num{1.19e-3} & \hfil\num{-8.50e-3}\\
         Second word position & \hfil x & \hfil\num{-1.02e-2} & \hfil\num{7.75e-3}\\
         Third word position & \hfil x & \hfil\num{9.71e-3} & \hfil\num{3.33e-3} \\
         Trial number & \hfil\num{-5.32e-4} & \hfil\num{-6.93e-4} & \hfil\num{-9.11e-4}\\
        \bottomrule
      \end{tabular}
      \caption{Average coefficients of linear regression modeling  participant reaction time. All the coefficients related to the interpretability methods are negative. It indicates a positive effect on reaction time. The value of the CLS-A parameter is lower than the others methods for experiment 1 and 3.}
      \label{tab:table_coef_regression}
\end{table}

\subsection{Modeling summary} \label{B}
\begin{table}[H]
\centering
\begin{tabular}{|c|l|c|c|}
\hline
\textbf{\begin{tabular}[c]{@{}c@{}}Task \\ type\end{tabular}} & \textbf{Target variable} & \textbf{Model}               & \textbf{\begin{tabular}[c]{@{}c@{}}Explanatory \\ variables\end{tabular}}                                                                                                                                                    \\ \hline
Regression                                                    & Reaction time            & Linear model                 & \begin{tabular}[c]{@{}c@{}}expected answer, classifier probability score, \\ review length, trial number, interpretability method, \\ relative positions of the first, \\ second and third most impacting words\end{tabular} \\ \hline
Classification                                                & Accurate                 & Explainable boosting machine & \begin{tabular}[c]{@{}c@{}}reaction time, classifier probability score, \\ review length, trial number, interpretability method, \\ relative position of the first most impacting word\end{tabular}                          \\ \hline
\end{tabular}
\caption{Linear regression and explainable boosting machine explanatory variables. The variables of the relative positions of the second and third most important words were used only for reaction time modeling in the first .}
      \label{tab:variables}
\end{table}

\subsection{Explainable boosting machine performance metrics} \label{D}
\begin{table}[H]
\centering
\begin{tabular}{@{}cccccc@{}}
\toprule
\textbf{Experiment}    & \textbf{Method} & \textbf{Accuracy} & \textbf{Precision Score} & \textbf{F1 Score} & \textbf{Recall Score} \\ \midrule
\multirow{4}{*}{Exp 1} & CLS-A           & 0.952             & 0.945                    & 0.953             & 0.961                 \\
                       & LIME            & 0.992             & 0.991                    & 0.992             & 0.993                 \\
                       & SHAP            & 0.961             & 0.955                    & 0.961             & 0.976                 \\
                       & Random          & 0.982             & 0.988                    & 0.982             & 0.976                 \\ \midrule
\multirow{4}{*}{Exp 2} & CLS-A           & 0.889             & 0.894                    & 0.888             & 0.884                 \\
                       & LIME            & 0.897             & 0.888                    & 0.898             & 0.909                 \\
                       & SHAP            & 0.913             & 0.917                    & 0.913             & 0.909                 \\
                       & Random          & 0.878             & 0.867                    & 0.880             & 0.894                 \\ \midrule
\multirow{4}{*}{Exp 3} & CLS-A           & 0.957             & 0.950                    & 0.957             & 0.965                 \\
                       & LIME            & 0.946             & 0.943                    & 0.946             & 0.949                 \\
                       & SHAP            & 0.920             & 0.916                    & 0.920             & 0.925                 \\
                       & Random          & 0.920             & 0.921                    & 0.920             & 0.919                 \\ \cmidrule(l){2-6} 
\end{tabular}
\caption{Average explainable boosting machine performance per experiment, per method.}
      \label{tab:ebm_scores}
\end{table}

\newpage
\subsection{EBM reponse curves} \label{B}

\begin{figure}[h]
\centerline{\includegraphics[width=1\linewidth, height=5cm]{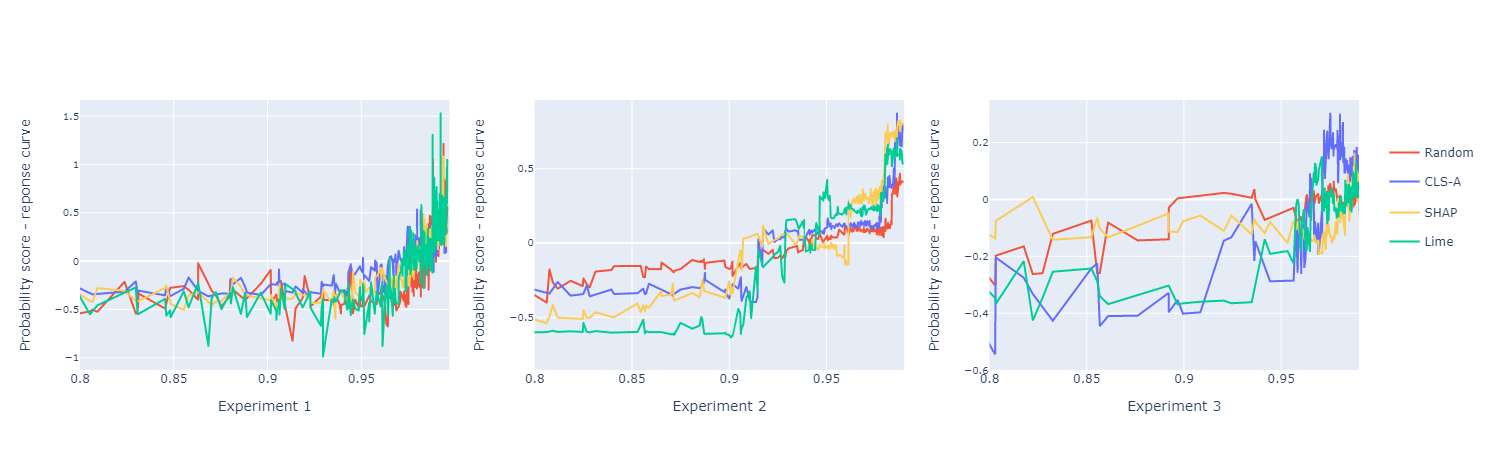}}
\caption{Explainable boosting machine response curves of probability score for all methods. The contribution of the probability score variable becomes higher for CLS-A, SHAP and LIME above a certain probability threshold compared to the random baseline for Experiment 2 and 3.}
\label{fig:EBM_proba_allmethods}
\end{figure}

\begin{figure}[h]
\centerline{\includegraphics[width=1\linewidth, height=5cm]{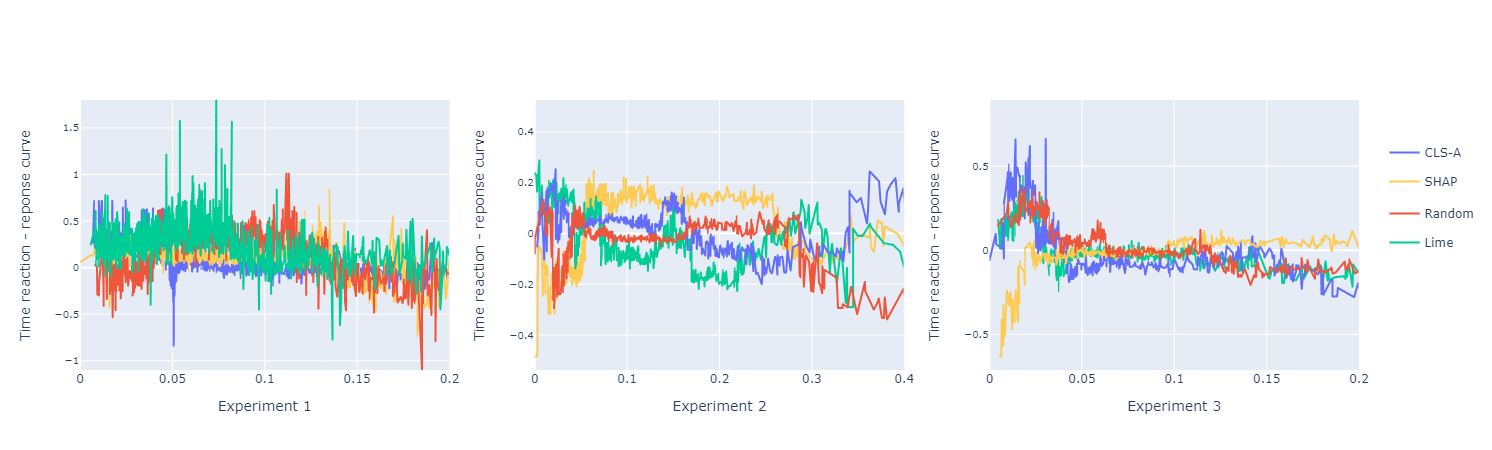}}
\caption{Explainable boosting machine response curves of time reaction for all methods. }
\label{fig:EBM_time_allmethods}
\end{figure}

\begin{figure}[h]
\centerline{\includegraphics[width=1\linewidth, height=5cm]{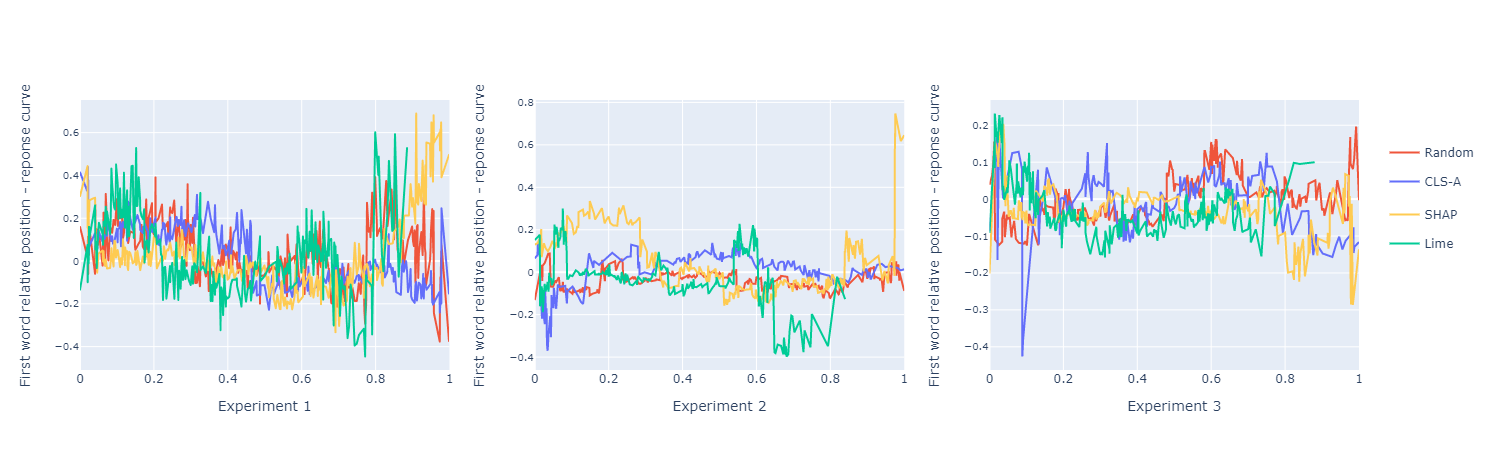}}
\caption{Explainable boosting machine response curves of first word relative position.}
\label{fig:EBM_first_word}
\end{figure}
\begin{figure}[h]
\centerline{\includegraphics[width=1\linewidth, height=5cm]{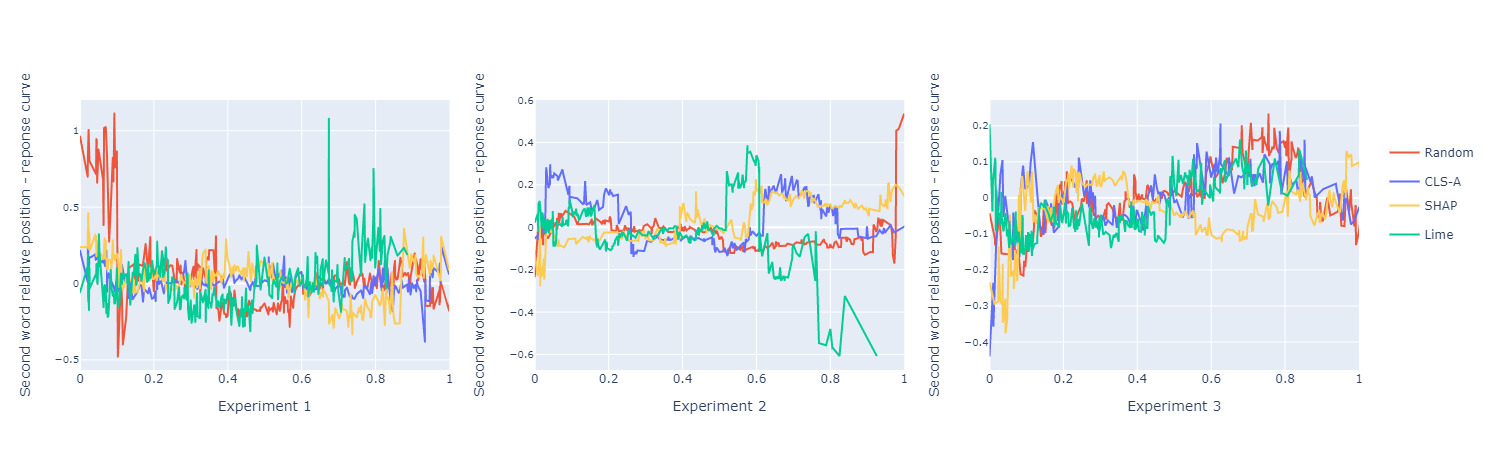}}
\caption{Explainable boosting machine response curves of second word relative position}
\label{fig:EBM_second_word}
\end{figure}
\begin{figure}[h]
\centerline{\includegraphics[width=1\linewidth, height=5cm]{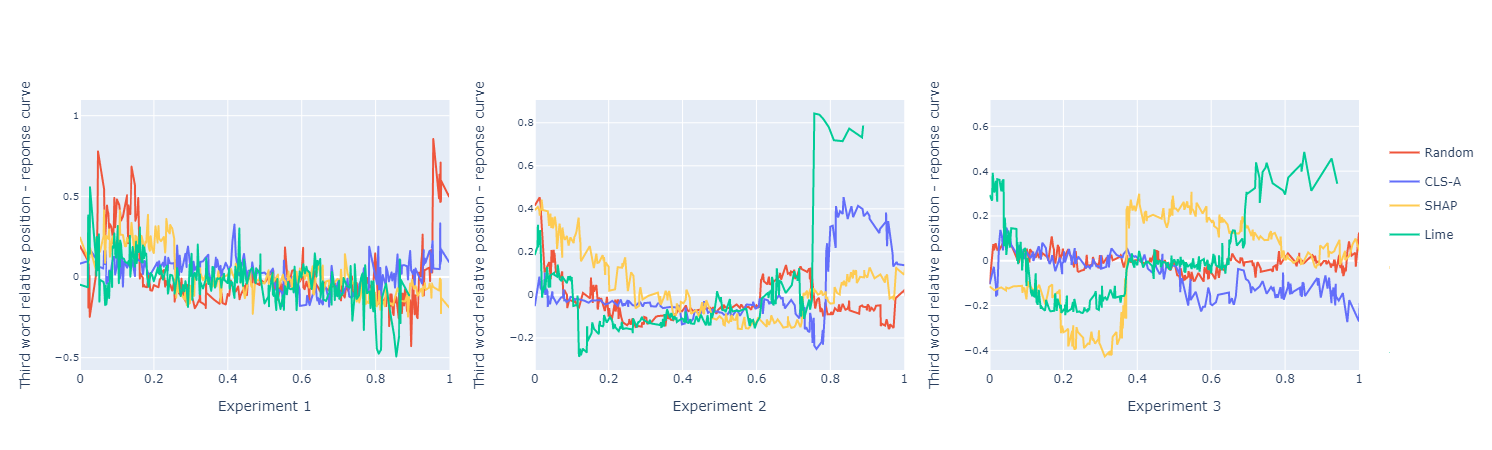}}
\caption{Explainable boosting machine response curves of third word relative position}
\label{fig:EBM_third_word}
\end{figure}
\begin{figure}[h]
\centerline{\includegraphics[width=1\linewidth, height=5cm]{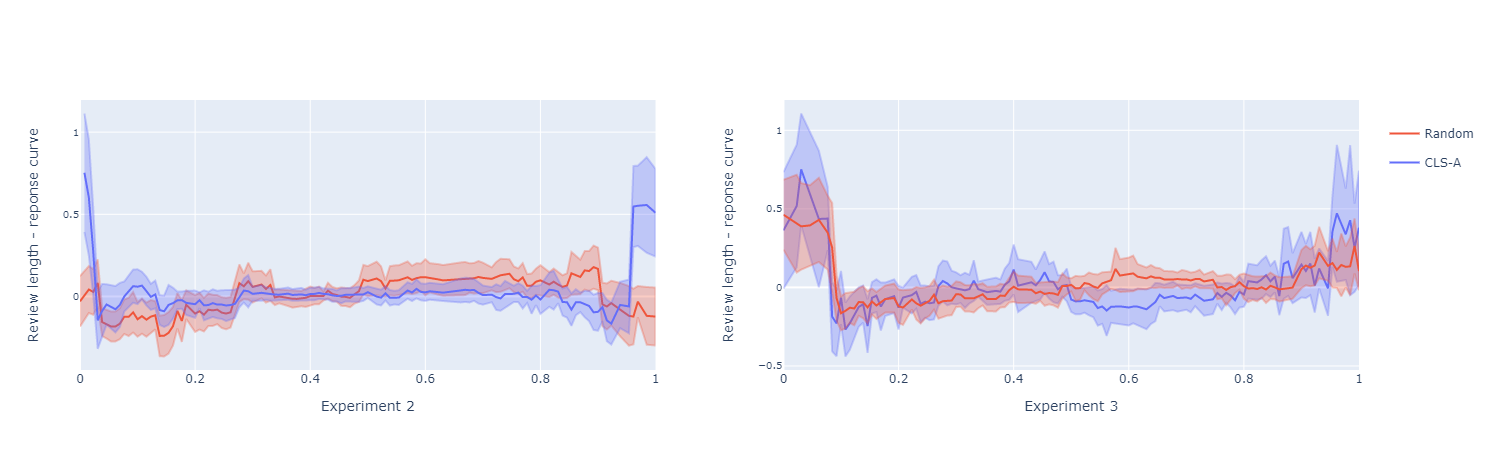}}
\caption{Explainable boosting machine response curves of review length. The contribution of the review length variable is higher for CLS-A for very short and very long reviews.}
\label{fig:EBM_rev_length_xp23}
\end{figure}


\end{document}